\documentclass[conference]{IEEEtran}
\ifCLASSINFOpdf
\else
\fi
\usepackage{times}
\usepackage{epsfig}
\usepackage{graphicx}
\usepackage{amsmath}
\usepackage{amssymb}
\usepackage{multirow}
\usepackage{placeins}
\usepackage{float}

\begin{document}

%
\title{The Impact of Explanations on AI Competency Prediction in VQA}


\author{\IEEEauthorblockN{Kamran Alipour\IEEEauthorrefmark{1}, Arijit Ray\IEEEauthorrefmark{2}, Xiao Lin\IEEEauthorrefmark{2}, Jurgen P. Schulze\IEEEauthorrefmark{1}, Yi Yao\IEEEauthorrefmark{2}, and Giedrius T. Burachas\IEEEauthorrefmark{2}}
\IEEEauthorblockA{\IEEEauthorrefmark{1}UC San Diego, La Jolla, CA.
\\\{kalipour, jschulze\}@eng.ucsd.edu}
\IEEEauthorblockA{\IEEEauthorrefmark{2}SRI International, Princeton, NJ.\\\{arijit.ray, xiao.lin, yi.yao, giedrius.burachas\}@sri.com}


}


%


\maketitle

\begin{abstract}
Explainability is one of the key elements for building trust in AI systems. Among numerous attempts to make AI explainable, quantifying the effect of explanations remains a challenge in conducting human-AI collaborative tasks. Aside from the ability to predict the overall behavior of AI, in many applications, users need to understand an AI agent’s competency in different aspects of the task domain. In this paper, we evaluate the impact of explanations on the user’s mental model of AI agent competency within the task of visual question answering (VQA). We quantify users’ understanding of competency, based on the correlation between the actual system performance and user rankings. We introduce an explainable VQA system that uses spatial and object features and is powered by the BERT language model. Each group of users sees only one kind of explanation to rank the competencies of the VQA model. The proposed model is evaluated through between-subject experiments to probe explanations' impact on the user’s perception of competency. The comparison between two VQA models shows BERT based explanations and the use of object features improve the user’s prediction of the model’s competencies.
\end{abstract}


%
\IEEEpeerreviewmaketitle

\section{Introduction}
Recent developments in the field of AI and specifically deep neural networks (DNN) have brought them into a broad range of applications. DNNs have automated a wide range of human activities resulting in reduced complexity of many tasks. Users of AI systems, though, need to maintain at least a minimal level of understanding and trust in the system, i.e., they need a proper mental model of the system’s internal operations for anticipating success and failure modes.\\
While accuracy is well-known as the primary metric for AI efficiency, it cannot guarantee a collaborative human-machine interaction in the absence of trust. If the users do not trust a model or a prediction, they will not use it \cite{ribeiro2016should}. This mistrust escalates in the presence of adversarial attacks where imperceptible changes to the input lead to wrong outputs and also the susceptibility of DNNs to non-intuitive errors.\\
Explainable AI aims to gain user's trust on two major steps of interpretability and explainability. Interpretable models provide a basic comprehension of their inner-processes through visual or textual cues. On a higher level, explainable models attempt to provide reason and causality behind their decisions\cite{gilpin2018explaining}.\\
The appearance of various methods of explanations call for a parallel effort to evaluate and quantify their efficiency. While previous works introduce nominal visualizations and textual justifications on the inner features of DNN models; yet it does not evaluate the impact of explanations on various aspects of user’s understanding and trust.\\
Evaluation techniques for explanations include automatic and human methods. Automatic approaches provide quantifiable measures over relevant benchmarks e. g. alignment with human attention datasets\cite{das2017human}, however they still cannot  propose a straight-forward metric for trust in actual human-machine task.\\
Furthermore, human-based approaches attempt to quantify explanation effectiveness through collecting user ratings \cite{lu2016hierarchical,chandrasekaran2017takes}. Despite their insightful results, these methods do not measure user's perception of AI competency in the whole domain.\\
Users can benefit from AI systems more efficiently if they are familiar with the AI agent’s competency in the operational domain. The competency of AI can be impacted by the biases in the training data or limited representation of crucial features. An explanation system that provides case-by-case reasoning for AI behavior does not automatically produce a higher view of competency. Particularly, deep learning models are notoriously opaque and difficult to interpret and often have unexpected failure modes, making it hard to build trust.\\
As prior research shows explanations improve user’s prediction of system accuracy \cite{Alipour}. Herein, we focus on the user’s mental model of an AI system in terms of competency understanding. Specifically, we evaluate the importance of explanations for helping users interpret how a VQA system performs on different types of questions. We model users’ learning process under two different explanation systems to identify the role of the attention-based explanations in user’s prediction of competency. For this purpose, we evaluate the impact of explanations on user learning rate and also their ultimate score on the task of competency prediction. 
\section{Related work}
\textbf{Visual question answering (VQA).}
Originally introduced by \cite{antol2015vqa}, the VQA problem involves the task of answering questions about the visual content of an image. The VQA task is specifically challenging due to the complex interplay between the language and visual modalities\cite{zhang2019interpretable}. Limited labeled data and the complex feature space complicate the process of developing VQA models. These challenges result in models with inconsistent outputs and serious logical contradictions\cite{ray2019sunny}. In such an environment, the choice of hyper-parameters and architectural designs can have drastic impacts on the performance of VQA models\cite{teney2018tips}.\\
A common approach to VQA is to use DNNs with attention layers that select specific regions of the image, guided by the question for inferring an answer\cite{DBLP:journals/corr/abs-1708-02711,DBLP:journals/corr/XuS15a,10.1007/978-3-319-46478-7_28,fukui2016multimodal}. Herein, we also study two attention based VQA models with different attention structures. As a baseline, we use a model based on Kazemi and Elqursh \cite{DBLP:journals/corr/KazemiE17} and Teney \emph{et al.}\cite{DBLP:journals/corr/abs-1708-02711} approaches. We propose a new VQA architecture by replacing the attention mechanism with a BERT model\cite{Devlin} in the baseline VQA model.\\
The previous work in VQA includes various attempts to optimize the attention mechanism. To improve the attention on the question, Lu \emph{et al.}\cite{DBLP:journals/corr/LuYBP16} utilize a co-attention model to
jointly reason about image and question on hierarchical levels. Anderson \emph{et al.}\cite{anderson2018bottom} propose a combined bottom-up and top-down attention mechanism to calculate attention at the level of objects. The model is further upgraded and fine-tuned to win the VQA Challenge 2018\cite{jiang2018pythia}.\\ 
Despite all the advancement in the overall accuracy of VQA models, their unbalanced performance in different aspects of the task is overtly noticeable. Some prior approaches address this issue by focusing on certain tasks such as reading text in images\cite{Singh_2019_CVPR} or counting objects\cite{zhang2018learning}. 
Other works introduce new datasets to reduce bias \cite{balanced_vqa_v2} or to enforce the logical consistency of model through \textit{visual commonsense reasoning (VCR)} for challenging questions\cite{zellers2019recognition}.\\

\textbf{Explainable AI (XAI).}
The ever increasing complexity of the modern AI machine demands a trustable source of explanation for all the AI users. Generating automated reasoning and explanations dates back to very early work in the AI field with direct applications from medicine  \cite{shortliffe1984model}and education \cite{lane2005explainable,van2004explainable}, to robotics \cite{lomas2012explaining}. In the field of computer vision, several explanation systems focus on the importance of image features in the decision-making process \cite{zeiler2014visualizing,hendricks2016generating,jiang2017learning,jiang2018pythia}.\\
AI explanations for the task of visual question answering usually include image and language attentions  \cite{lu2016hierarchical,DBLP:journals/corr/KazemiE17}. Besides saliency/attention maps, other efforts investigated different explanation modes like layered attentions \cite{yang2016stacked}, bounding boxes around important regions \cite{anne2018grounding}, textual justifications \cite{shortliffe1984model,huk2018multimodal} or a combination of these modes \cite{Alipour}.\\
We propose an explainable VQA system which produces justifications for system answer in the form of an attention map. Unlike previous post-hoc saliency approaches such as GradCAM\cite{selvaraju2017grad}, our method seeks causal explanations by providing attentions as an inherent step of answer inference. Our proposed model uses visual features on both spatial and object level. For better performance in VQA task, the proposed model utilizes BERT language model to process question features along with the visual features.\\

\textbf{Explanation evaluation.} As the AI machines enter the daily life of people, a new interest has surged among the AI community to make AI algorithms more understandable to the lay users without the technical background\cite{10.1145/3377325.3377501}. In this work, we choose the subjects for explanation evaluation from a group of individuals with minimum knowledge about AI and deep neural networks.\\
Evaluating the impact of explanations on user mental model and human-machine performance is widely discussed in the XAI literature. Some of the earlier works take on quantifying the efficacy of explanations through user studies to assess the role of explanations in building a better mental model of AI systems for their human users.\\
Some of the previous studies introduced metrics to measure trust with users \cite{cosley2003seeing,ribeiro2016should}, or the role of explanations to achieve a goal \cite{kulesza2012tell,narayanan2018humans,ray2019lucid}. Dodge \emph{et al.} investigated the fairness aspect of explanations through emperical studies\cite{10.1145/3301275.3302310}. Lai and Tan \cite{lai2019human} assessed  the role of explanations in user success within a spectrum from human agency to full machine agency. Lage \emph{et al.} proposed a method to evaluate and optimize human-interpretability of explanations based on measures such as size and repeated terms in explanations\cite{lage2019evaluation}. Other approaches measured the effectiveness of explanations in improving the predictability of a VQA model \cite{chandrasekaran2018explanations,Alipour}.\\
In this work, we conduct a user study to investigate the impact of explanations on the user’s mental model of system competency. Within the study, subjects attempt to rank system performance among different types of input questions. The results indicate a positive influence on the accuracy of the user’s mental model in the presence of explanations. We detail the overall and temporal effect of explanations on the user’s interpretation in two explainable VQA models.\\
\section{Methods}
Our approach aims at evaluating the role of attentional explanations in the user's mental model of AI competency. To accomplish this task, we compare two explainable VQA models and test them through user studies.\\
In this section, we cover the architecture details for these VQA models and the differences in their attention mechanisms. The section later follows with sample cases from both explanation models and the differences between them.\\
\begin{figure*}[ht]
\centering
\includegraphics[width=2.1\columnwidth]{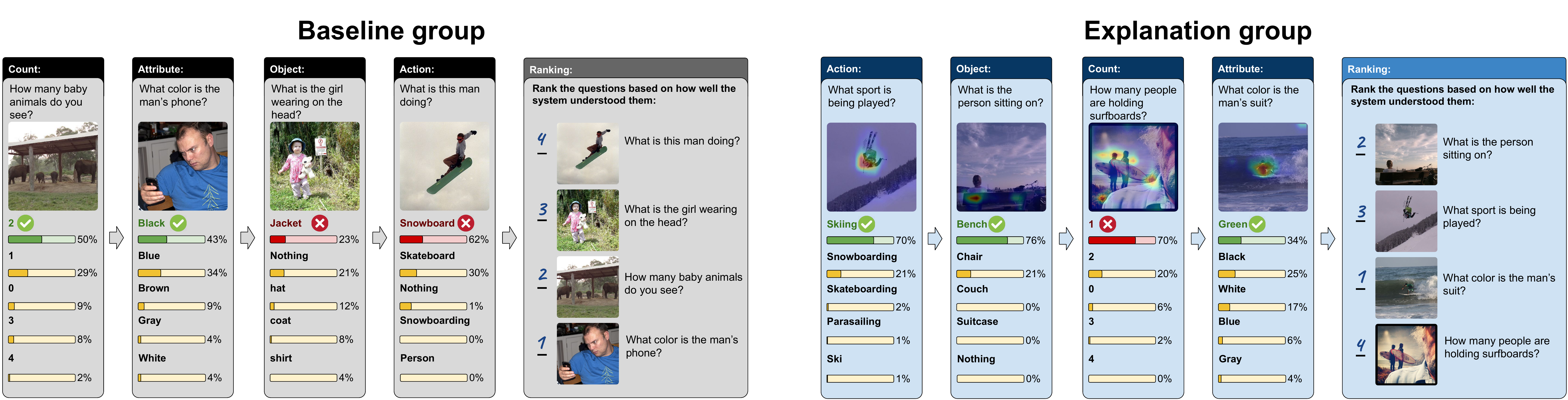}
    \caption{The workflow for user study groups: Left shows the baseline group where the users only view the top five answers from the model along with the probability of the answers. As shown on the right, users inside the explanation group, also view the attention maps generated by the model. Each group views blocks of trials. At the end of each block, users are asked to rank the question-images based on how well they seem to be \emph{understood} by the model.}
     \label{fig:competency_task_flowchart}
\end{figure*}
\subsection{Explainable VQA (XVQA) models}
\begin{figure}[ht]
  \centering
  \includegraphics[width=1.0\columnwidth]{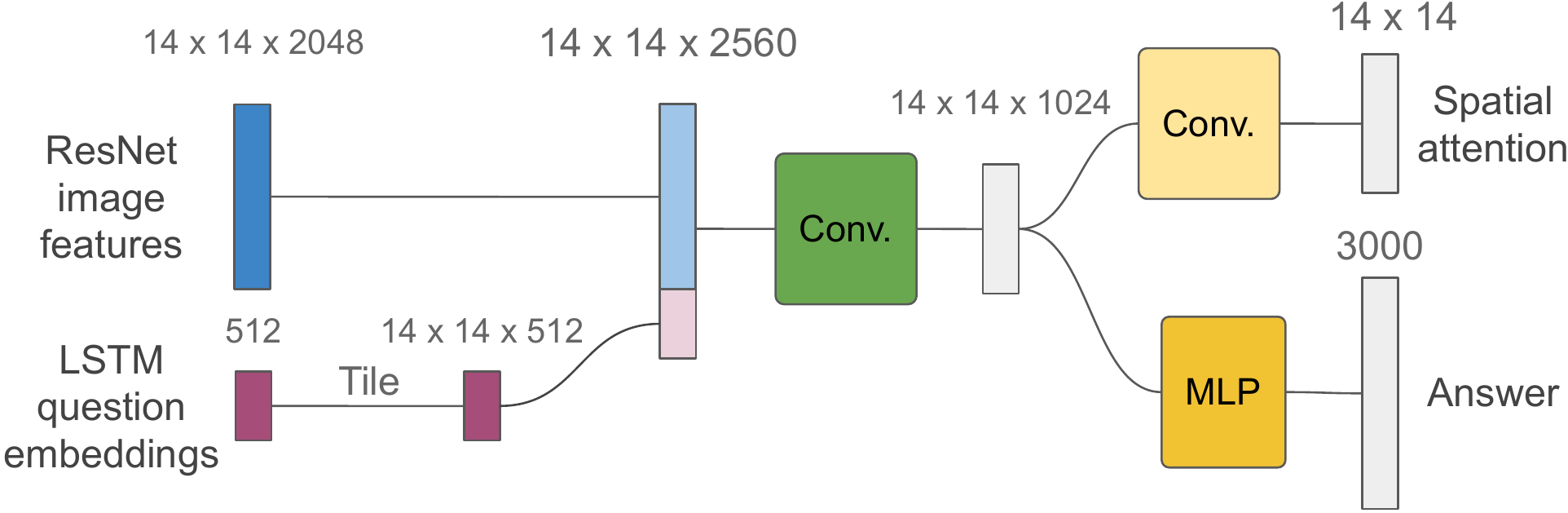}
    \caption{The architecture of explainable SVQA model.}
     \label{fig:svqa}
\end{figure}

Our work compares two VQA agents: spatial attention VQA (SVQA) and spatial-object attention BERT VQA (SOBERT). Both agents are trained on VQA 2.0 dataset. SVQA is based on a 2017 SOTA VQA model with a ResNet \cite{szegedy2017inception} image encoder (figure \ref{fig:svqa}). The agent uses an attention mechanism to select visual features generated by an image encoder and an answer classifier that predicts an answer from 3000 candidates.\\
As shown in figure \ref{fig:svqa}, SVQA takes as input a $224 \times 224$ RGB image and question with at most 15 words. A ResNet subnet encodes the image into a $14 \times 14 \times 2048$ feature representation. An LSTM model (GloVe \cite{pennington2014glove}) encodes the input question word embeddings into a feature vector of 512 dimensions. \\
The attention layer in the SVQA model transfers the question and image features to a set of attention weights on the image features. The model convolves the concatenation of weighted image features and question features to produce the attention layer with $14 \times 14 \times 1024$ dimensions. The model predicts the probability of the final answer from a set of 3000 answer choices using a multilayer perceptron (MLP). The attention layer also goes through a convolution block to generate the spatial attention map.\\

\begin{figure}[ht]
  \centering
  \includegraphics[width=1.0\columnwidth]{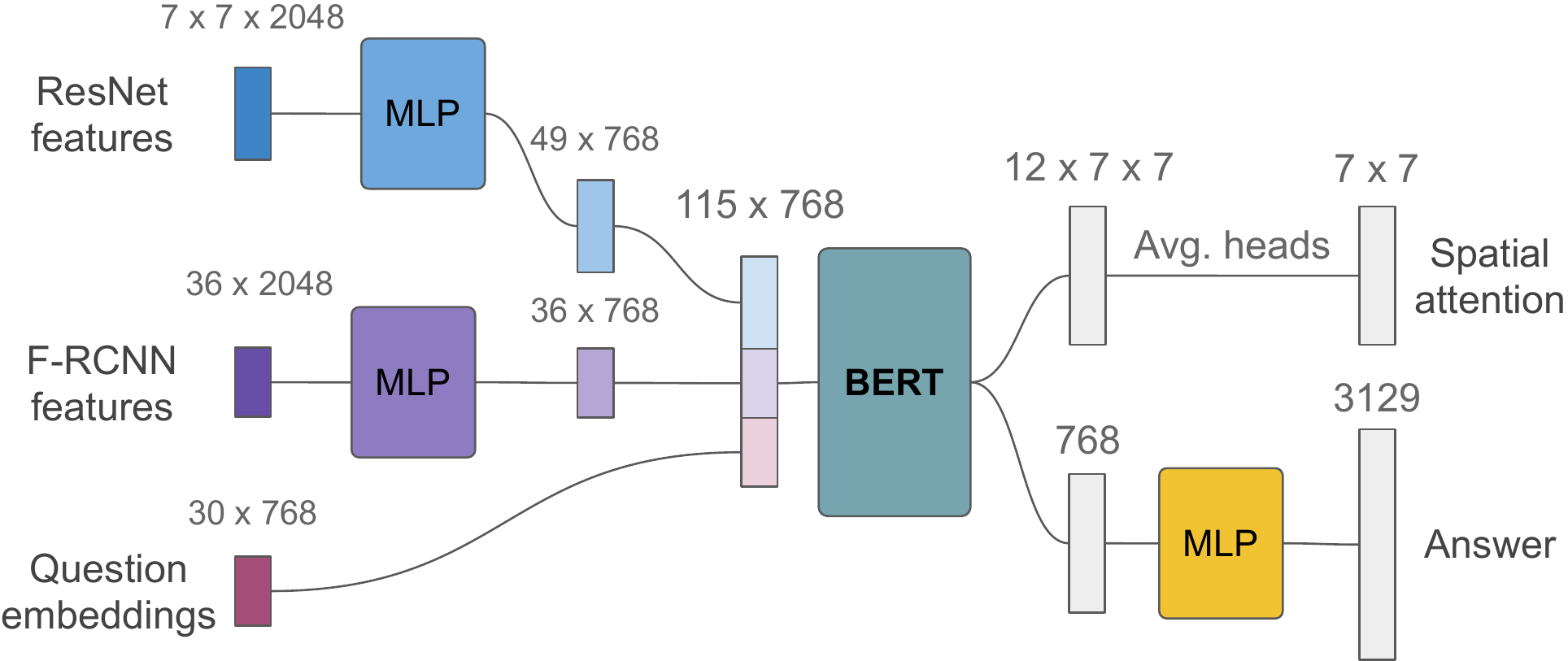}
    \caption{The architecture of the explainable SOBERT model. This model passes the combination of visual features from ResNet and FRCNN and question embeddings into a BERT model to produce answers and spatial attention.}
     \label{fig:sobert}
\end{figure}

On the other hand, the SOBERT agent uses a combination of visual embeddings of the image from ResNet and Faster RCNN (FRCNN)\cite{Ren} alongside question embeddings (figure \ref{fig:sobert}). SOBERT accepts questions with a maximum length of 30 words and the input question embeddings contain the location and token information of words. The location features are encoded in both ResNet and question embeddings.\\
SOBERT agent uses a BERT model with 4 layers and 12 attention heads. BERT transfers the hidden features ($115 \times 768$) into spatial attention heads ($12 \times 7 \times 7$) and output layer. An MLP maps the output layer to the final answer prediction out of 3129 candidates.\\
Based on their training process and their characteristics, VQA agents can reach certain levels of accuracy in each type of question. For our tests, we limit the cases into a subset of VQA 2.0 validation set with questions about action, attribute, object, and count. We classified the question using a set of automated methods including word matching in questions and also their answers.\\ 
Questions about activity inside an image are labeled as "Action". Questions about objects inside the image are labeled as "Object". Questions that are specific about attributes of entities in the image (e.g. color) are labeled as "Attribute". Finally, questions about counting entities on the image are categorized as "Count". Table \ref{table:sys_accs} shows the accuracy of SVQA and SOBERT agents in these four categories. The accuracy of models are computed over the four categories within VQA validation dataset.\\
As numbers in table \ref{table:sys_accs} show, the two models pose a similar ranking between the four categories of questions, while the SOBERT model can reach a higher accuracy in all of them compared to the SVQA model.


\begin{table}[ht]
  \centering
  \renewcommand{\arraystretch}{2}
  \begin{tabular}{c c c c c}
    {} & {Action} 
    & {Attribute}
    & {Object}
    & {Count}\\
    \hline
    {SVQA} & {$81.21\%$} & {$70.83\%$} & {$64.46\%$} & {$45.78\%$}\\
    \hline
    {SOBERT} & {$\textbf{88.35\%}$} & {$\textbf{86.63\%}$} & {$\textbf{71.84\%}$} & {$\textbf{60.14\%}$} \\
    \hline
    \\
  \end{tabular}
  \renewcommand{\arraystretch}{1}
  \caption{The accuracy of VQA agents in four selected categories of question.}\label{table:sys_accs}
\end{table}

\begin{figure*}[ht]
\centering
\begin{tabular}{cc}
\begin{tabular}{cc}
\multicolumn{2}{c}{\textbf{Action}: Is the animal sitting or standing?} \\ 
SVQA & SOBERT \\
\includegraphics[width=.49\columnwidth]{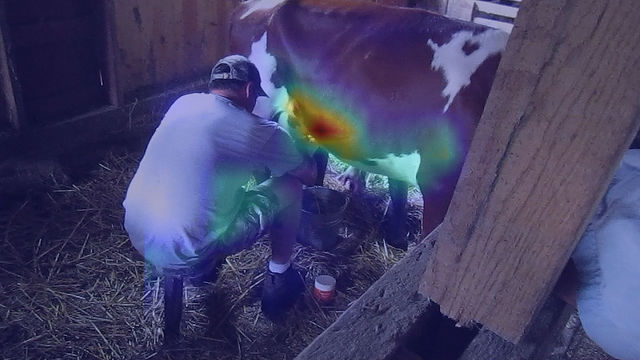} & 
\includegraphics[width=.49\columnwidth]{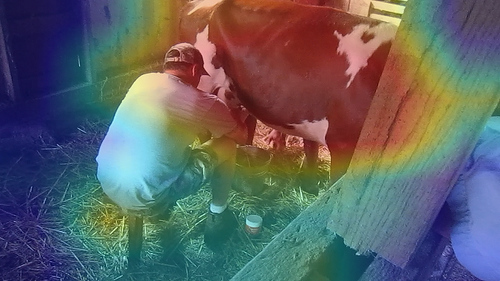}\\
\textbf{Ans:} Sitting & \textbf{Ans:} Standing \\
\end{tabular} &
\begin{tabular}{cc}
\multicolumn{2}{c}{\textbf{Object}: What is on the shelf?} \\ SVQA & SOBERT \\
\includegraphics[width=.49\columnwidth]{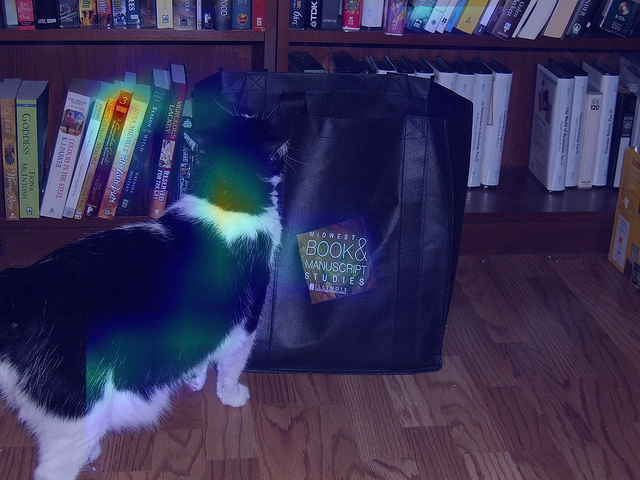} & \includegraphics[width=.49\columnwidth]{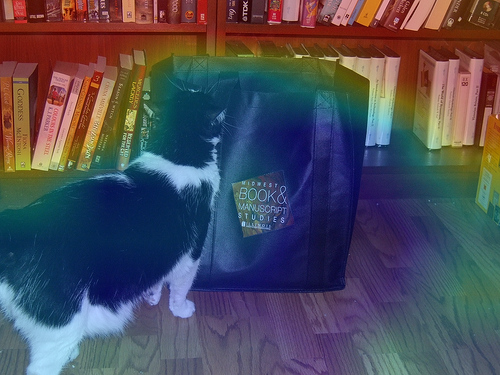} \\
\textbf{Ans:} Cat & \textbf{Ans:} Books \\
\end{tabular} \\
\begin{tabular}{cc}
\multicolumn{2}{c}{\textbf{Attribute}: What color is the cat?} \\
SVQA & SOBERT \\
\includegraphics[width=.49\columnwidth]{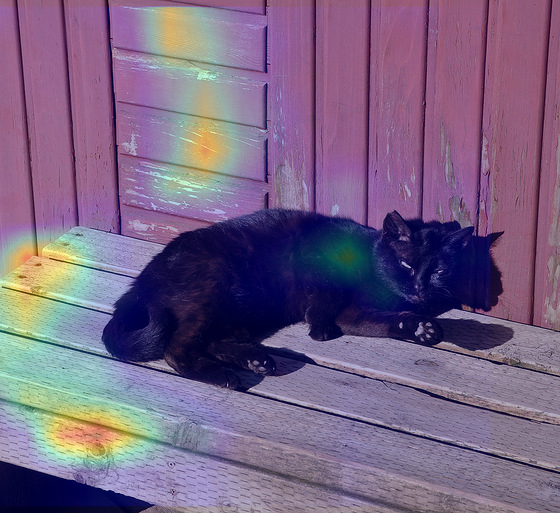} & \includegraphics[width=.49\columnwidth]{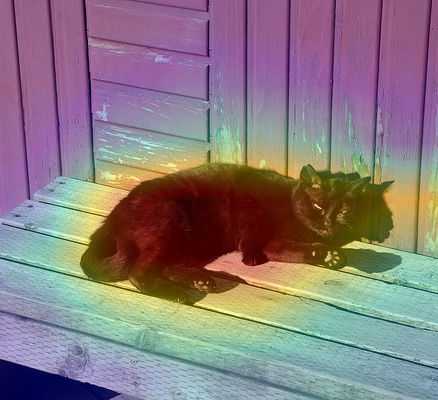} \\
\textbf{Ans:} Brown & \textbf{Ans:} Black \\
\end{tabular} &
\begin{tabular}{cc}
\multicolumn{2}{c}{\textbf{Count}: How many  zebras  on there?} \\
SVQA & SOBERT \\
\includegraphics[width=.49\columnwidth]{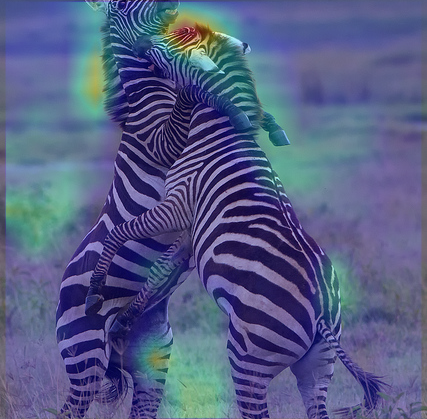} & \includegraphics[width=.49\columnwidth]{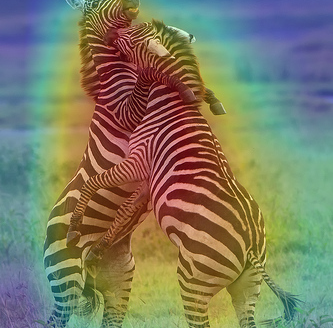} \\
\textbf{Ans:} 1 & \textbf{Ans:} 2 \\
\end{tabular}
\end{tabular}
\caption{Attention maps generated by the AI agents for questions in different question type categories. As illustrated in the results, the SOBERT model produces attention maps with more focus on the areas related to the question.}
\label{fig:exp_examples}
\end{figure*}

\subsection{Explanations}
The VQA agents can produce a spatial attention map to visualize the areas of focus while producing the answer. SVQA model convolves the attention tensor into a $14 \times 14$ spatial map. In the SOBERT model, the attention tensor is averaged over the 12 attention heads into a $7 \times 7$ spatial attention map.\\
The attention maps generated by the VQA agents provide a causal explanation to the users as they illustrate AI spatial/object attentions as an inherent step in answer inference. Both models use spatial features from the images while gaining a general representation of image content.
 SOBERT model also incorporates object-level  F-RCNN features into the process.\\
 One major impact of including object-level attention emerges in the attention map outputs of the model. As can be seen in figure \ref{fig:exp_examples}, the attentions from the SOBERT model cover broader areas that are associated with objects on the scene. Also, the averaging layer that generates attention produces smooth attention distributions in the SOBERT model compared to more localized and scattered attention in SVQA.\\
\begin{table*}[h]
 \centering
    \renewcommand{\arraystretch}{1.5}
        \begin{tabular}{llll}
            \hline
    {Model} & {Condition} & 
    {Final ranking corr.} &
    {Max. user learning rate (corr. / blocks)} \\ \hline
        \multirow{2}{*}{SVQA} 
        & {Baseline} & \multicolumn{1}{c}{$0.757$} & \multicolumn{1}{c}{$0.0105$} \\
        \cline{2-4}
         & {Explanation} & \multicolumn{1}{c}{$\textbf{0.805}$} & \multicolumn{1}{c}{$\textbf{0.0769}$} \\
        \hline
       \multirow{2}{*}{SOBERT} 
        & {Baseline} & \multicolumn{1}{c}{$0.611$} & \multicolumn{1}{c}{$0.0253$} \\
        \cline{2-4}
         & {Explanation} & \multicolumn{1}{c}{$\textbf{0.921}$} & \multicolumn{1}{c}{$\textbf{0.0468}$} \\
        \hline
        \\
        \end{tabular}
        \renewcommand{\arraystretch}{1}
    
    \caption{
  The maximum learning rate of users and the final value of correlation in competency ranking task. Both explanation models show an improvement in early learning rates. While explanation from the SOBERT model does increase the learning rate as much as SVQA, however, SOBERT reaches a relatively higher final learning rate. }
    \label{table:learningrates}
\end{table*}
\section{Experiments}
\begin{figure}[ht]
\centering
\begin{tabular}{@{}c@{}}
  \centering
  \includegraphics[width=.99\columnwidth]{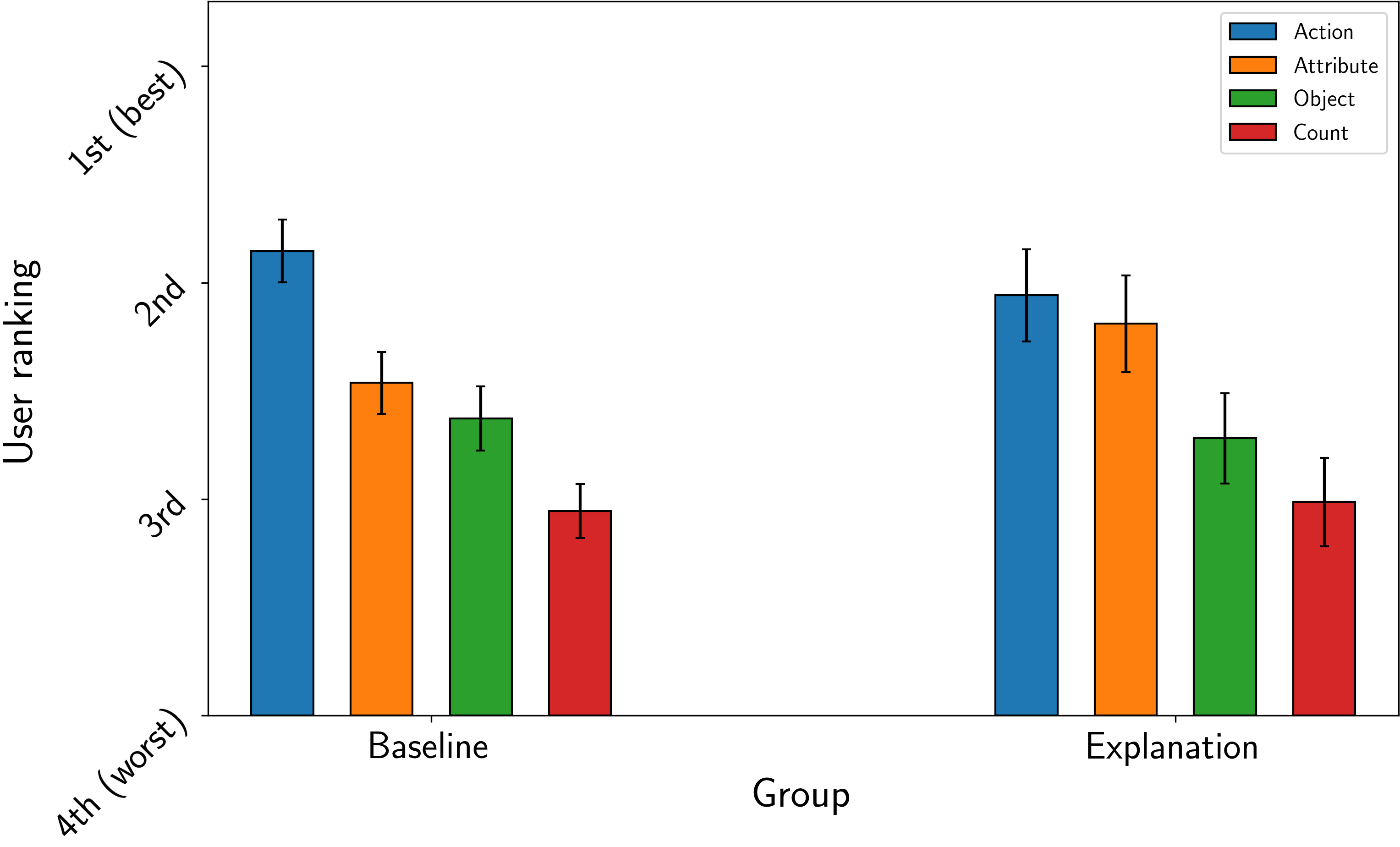}
 \end{tabular}
\begin{tabular}{@{}c@{}}
  \centering
  \includegraphics[width=.99\columnwidth]{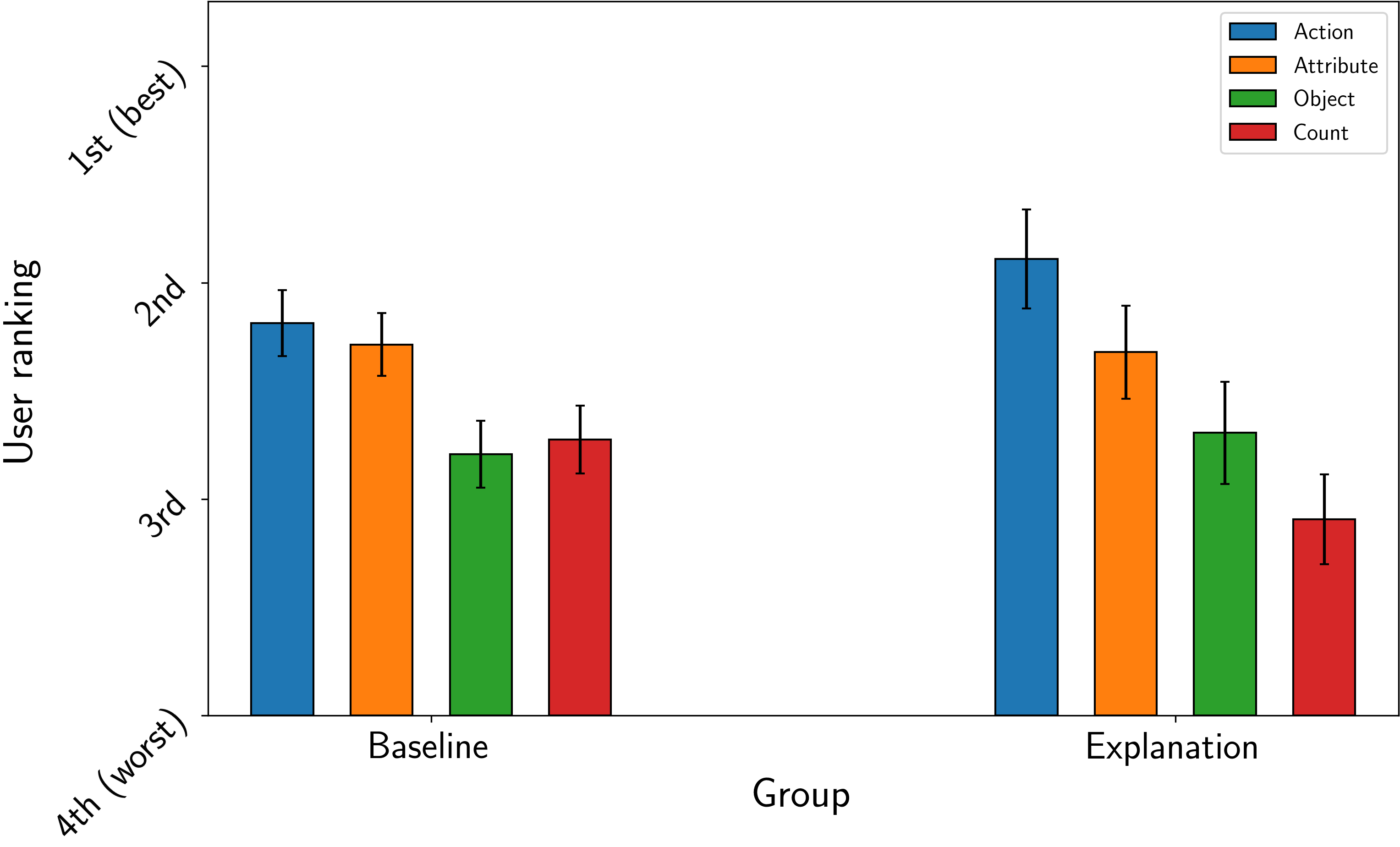}
 \end{tabular}
    \caption{The average of all rankings entered by the subjects at the end of every block of trials (Top: SVQA model, Bottom: SOBERT model).}
     \label{fig:ranking_competency}
\end{figure}

We designed an interface for an in-person user study to evaluate the impact of explanations on the user’s understanding of AI agent competency among different question types. At the introductory section of each study session, subjects are reminded that the model competency and accuracy of the AI model is unknown to minimize their prior knowledge and judgment of the AI agent competency.\\
In this user study, subjects go through a set of trial blocks where the AI agent answers questions about images. Each block consists of four trials with one image-question of each type: object, attribute, action, and count. On each trial, subjects first see the input image and question and then they proceed to see the outputs of the AI agents.\\
For each model, the study is divided into two groups of baseline and explanation. Each study group contains 10 subjects and each subject goes through 100 trials (25 blocks). In all groups, users see the agent’s top five answers, their probabilities, and agents Shannon's confidence in each trial. In the explanation group, subjects first view the attention map from the model and then see the top answers and confidence value. Subjects are asked to rank the helpfulness of attention maps on understanding AI's performance on that trial.\\
At the end of each block, subjects rank the trials within the block based on system performance in each question type. Comparing question type rankings from subjects between baseline and explanations measures the explanation on subjects’ opinions of system competency (figure \ref{fig:ranking_competency}).\\
In each block of trials, four question-images show up in random order. The AI agent's success ratio in each block is also random. Among the baseline group, users can rely on the top answers and their probabilities to understand system performance on that question and image. On the other hand, subjects from the explanation groups have the extra information provided by the attention maps (figure \ref{fig:competency_task_flowchart}).
\subsection{Explanation helpfulness}
\begin{figure}[ht]
\centering
\includegraphics[width=.99\columnwidth]{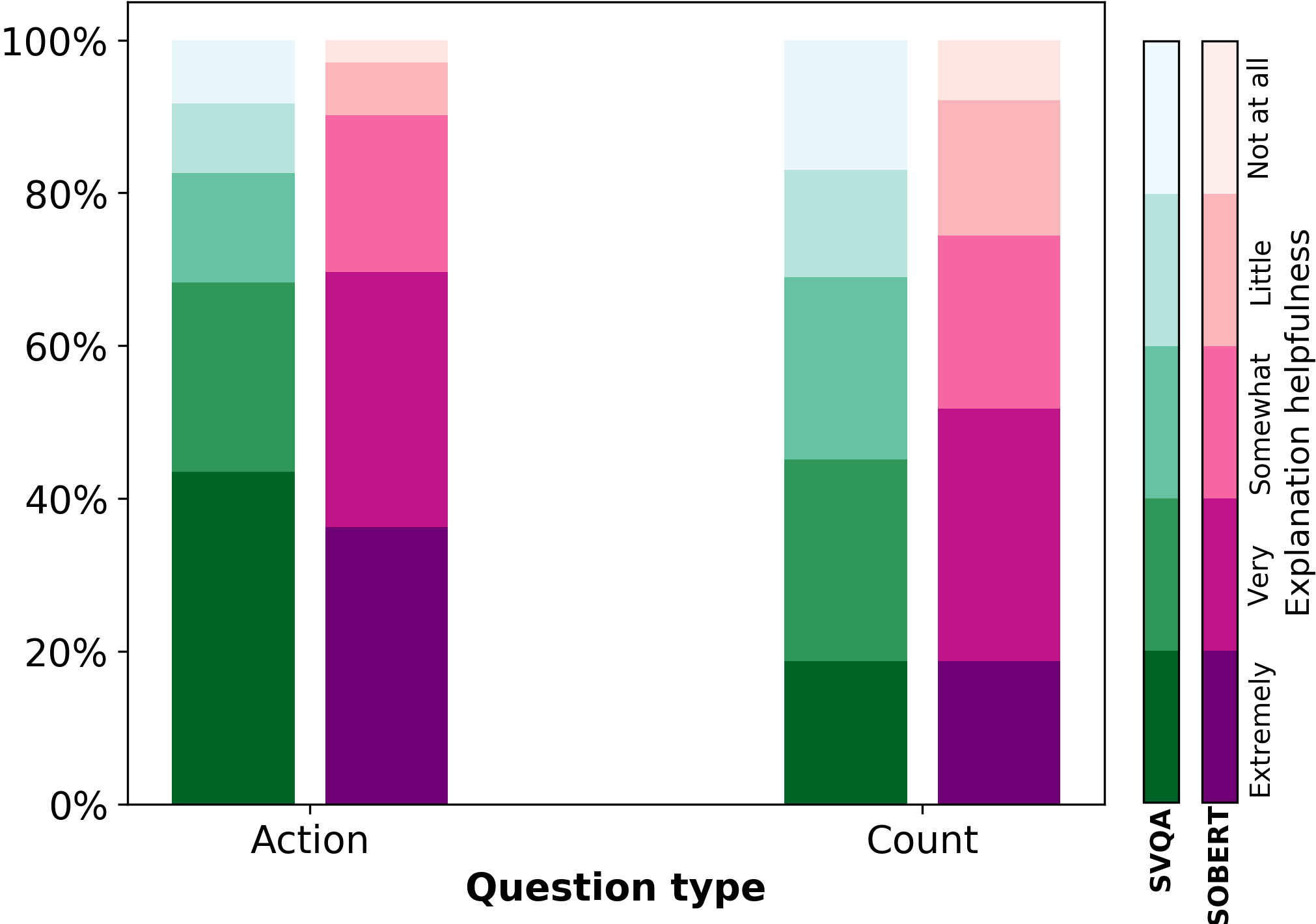}
    \caption{Histogram of ratings of how “helpful” explanations are for the subjects. These helpfulness ratings are given by the subjects as they view the explanations and before they see the system top 5 answers. So these ratings are not confounded by the accuracy of the AI.}
     \label{fig:helpfulness_histogram}
\end{figure}

In the explanation group, subjects view the attention explanations before they see the final answers and accuracy of AI. At this stage, subjects rate the explanations based on their helpfulness towards understanding AI's performance.\\ 
The helpfulness rankings are specifically interesting for action and count question types within which the VQA agents show their highest and lowest competencies. The helpfulness rankings of within these categories on SOBERT explanations show an increase compared to SVQA (figure \ref{fig:helpfulness_histogram}). While subjects rank 17\% of SVQA explanations are ranked as not helpful in count questions, this number is reduced to 7\% by SOBERT explanations. In action questions, SOBERT also reduces the unhelpful explanations from 8\% to 3\%.
\subsection{Competency ranking}
We assess the accuracy of subjects ranking by measuring the correlation between that and the ground truth competency ranking of AI agents (figure \ref{table:sys_accs}) and the collected rankings at the end of each block. Figure 6 illustrates this correlation in the starting and finishing blocks of each study group. The start and finish values of correlation are the average of 1-5 and 20-25 blocks respectively.\\

\begin{figure*}[ht]
\centering
\begin{tabular}{@{}c@{}}
  \centering
  \includegraphics[width=.99\columnwidth]{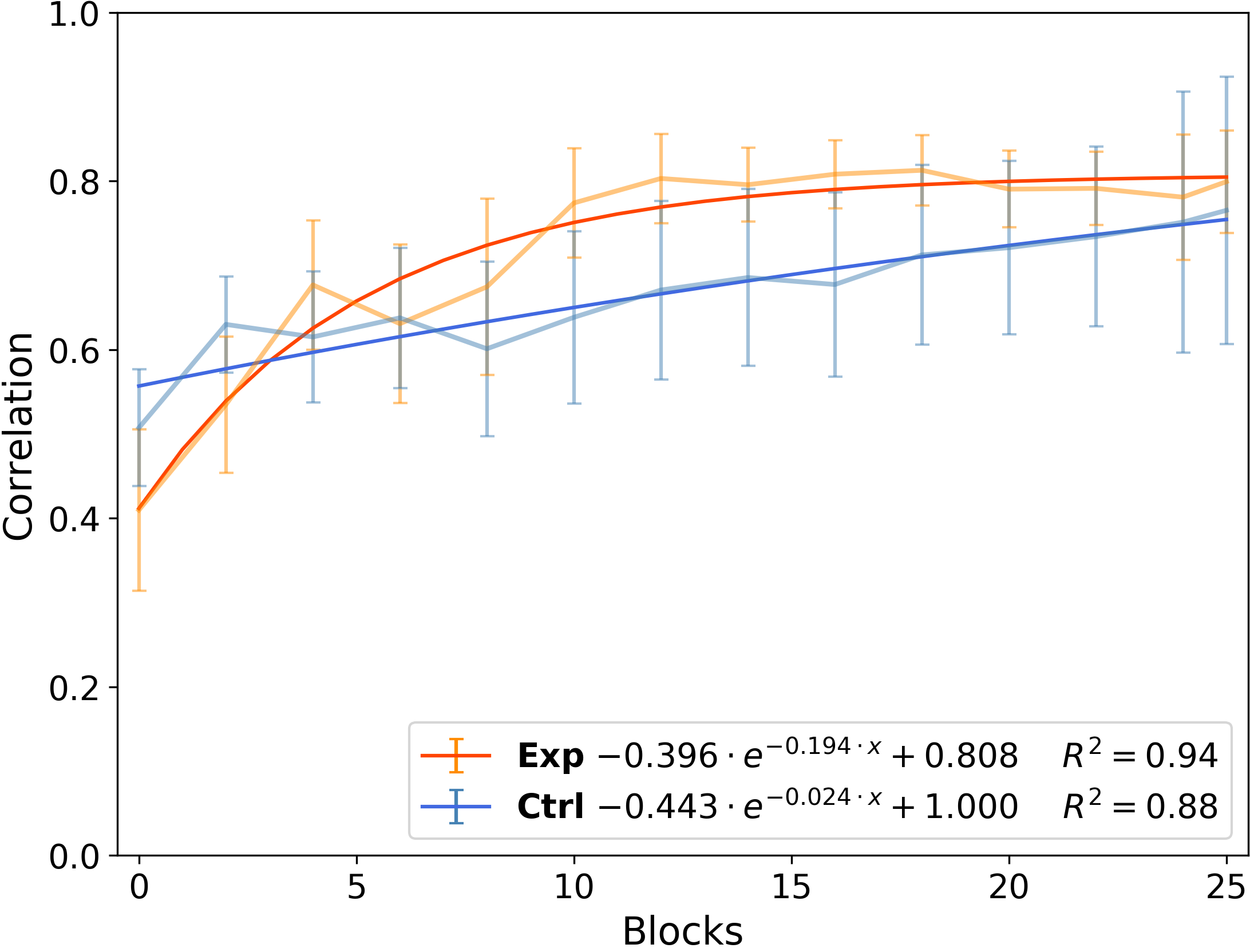}
 \end{tabular}
\begin{tabular}{@{}c@{}}
  \centering
  \includegraphics[width=.99\columnwidth]{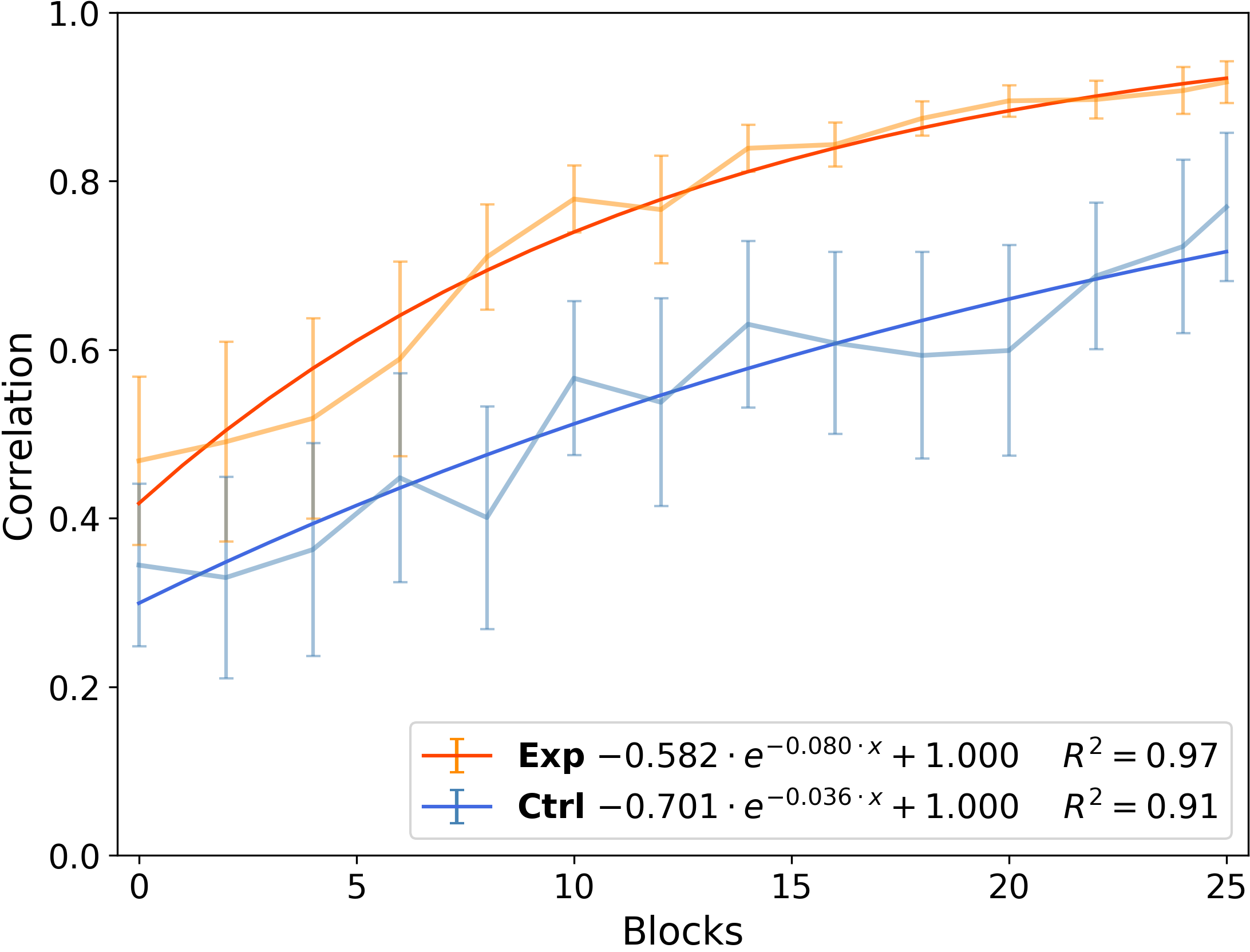}
 \end{tabular}
    \caption{Temporal impact of attention maps on user rankings. Left: the growth of correlation in baseline and explanation groups is compared between baseline (blue) and explanation (orange) groups for two models SVQA (left) and SOBERT (right). T-test p-values for SVQA and SOBERT data are $0.07$ and $3.7e-8$ respectively.}
     \label{fig:competency_progress}
\end{figure*}
\begin{figure}[ht]
\centering
\begin{tabular}{@{}c@{}}
  \centering
  \includegraphics[width=.49\columnwidth]{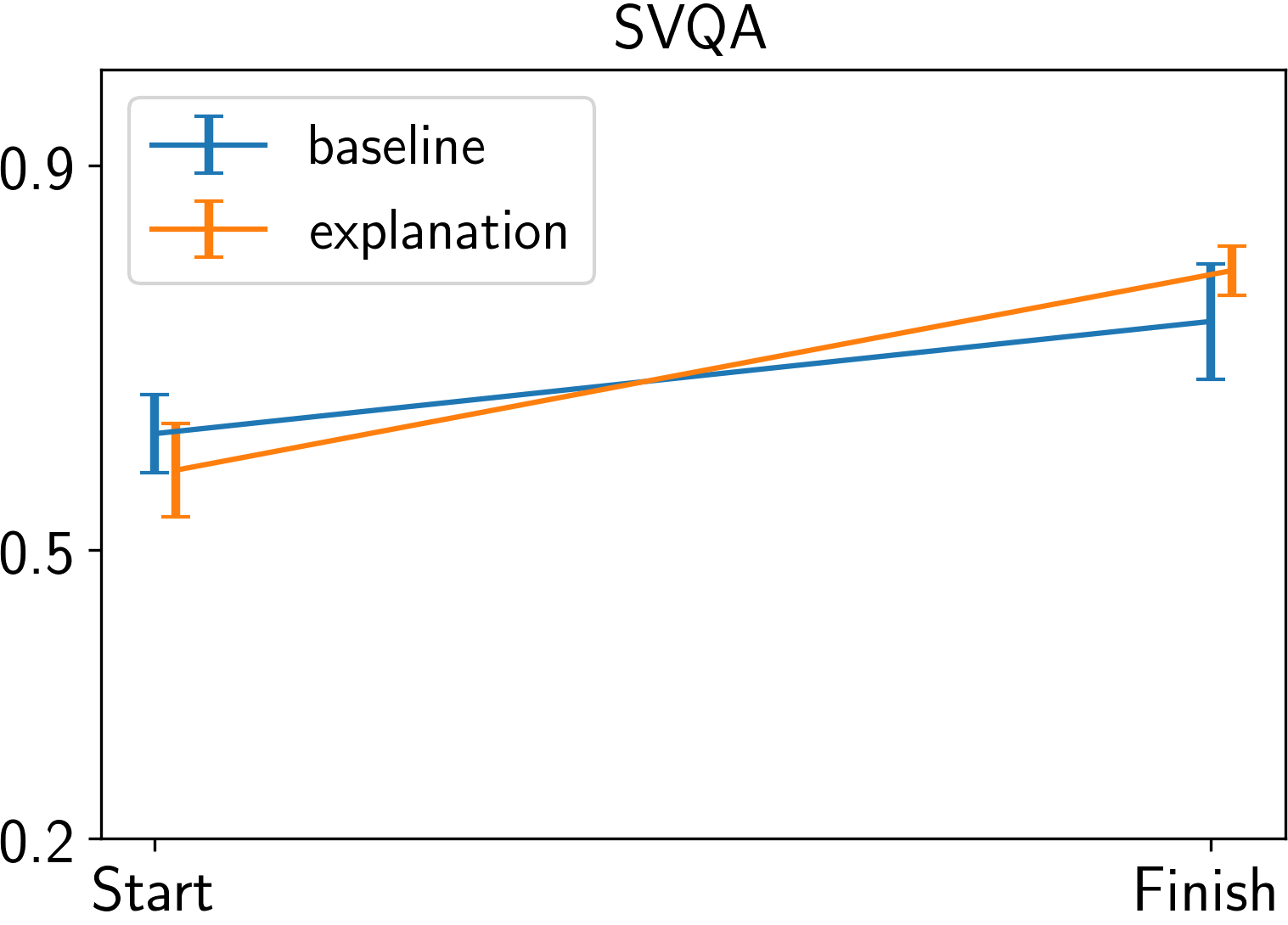}
 \end{tabular}
\begin{tabular}{@{}c@{}}
  \centering
  \includegraphics[width=.49\columnwidth]{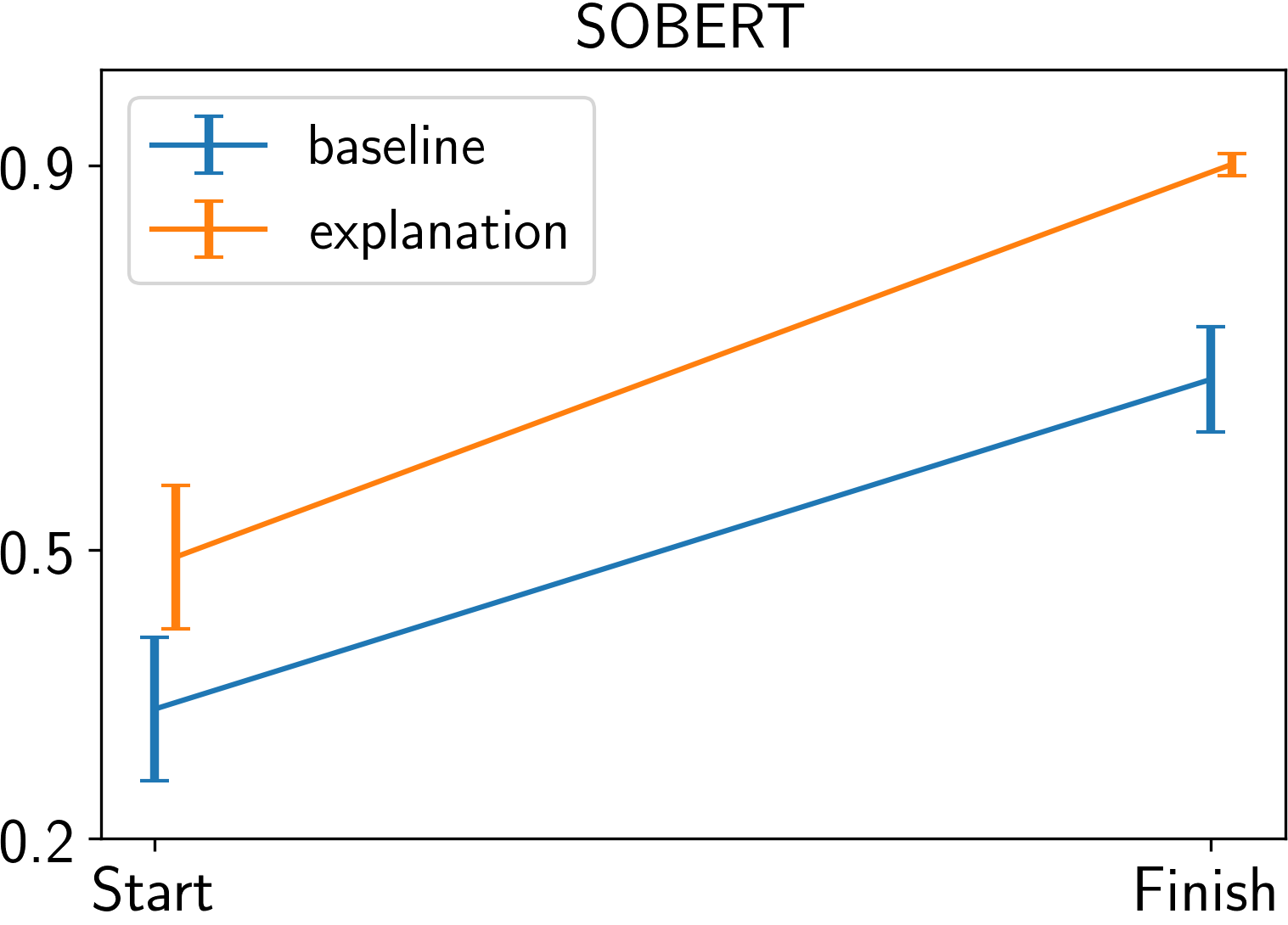}
 \end{tabular}
    \caption{The overall correlation between the user’s rankings and the system’s actual competencies. Comparing the results from the SVQA (left) model and our SOBERT (right) model suggests a better improvement of correlations in the presence of SOBERT attention maps.}
     \label{fig:overall_ranks}
\end{figure}

Overall, the ranking correlation shows an increase in both models with a slightly higher slope in the presence of explanations (table \ref{fig:overall_ranks}). To better picture the temporal impact of explanations on the users’ mental model, figure \ref{fig:competency_progress} presents the progress of ranking correlation throughout the study. In the early blocks for both models, the explanation groups increase their ranking correlation with a higher rate than baseline.
\subsection{Competency learning curves}
We also investigate the temporal pattern of temporal ranking correlation by fitting curves into the data in baseline and explanation groups. This problem, in general, can be viewed as modeling user learning curve for a certain task.\\
The modeling user learning curve is widely discussed in cognitive science. In previous works, researchers analytically derived exponential learning equations to describe user improvement in the task\cite{heathcote2000power,ritter2001learning}. The assumption of a monotonically decreasing improvement is the main foundation beneath the exponential learning curves.\\
Here in the context of learning AI competency rankings, subjects start the study with no prior knowledge of the AI agent's rankings. Also, the correlation metric cannot exceed the value of 1.0.\\
Considering these similarities to the general learning model, we also considered an exponential curve with an upper bound as blocks grow to infinity. With this analogy, we considered the following curve to fit the ranking correlation trends:
\begin{center}
$c = \alpha \cdot e^{-\beta \cdot b} + \delta$
\end{center}

where $b$ and $c$ are the block count and ranking correlation respectively. In this setting, the ranking correlation approaches $\delta$ as the subjects continue the study. The value of $\delta$ is penalized for curves fitting to satisfy the condition $\delta \leq 1.0$.\\
The slope of the fit curves in figure \ref{fig:competency_progress} represents the growth rate of ranking correlations with respect to the number of blocks. Higher rates of correlation growth show faster learning by the subjects. To compare the learning rates, we consider the maximum slope of each curve (table \ref{table:learningrates}).\\
The results indicate a higher rate of learning for users in the presence of an explanation. The explanation from the SVQA agent causes a higher increase in the learning rate compared to SOBERT. However, the ultimate value of ranking correlation in the SVQA model is bound to $\delta=0.808$ while the SOBERT model approaches the maximum correlation at $\delta=1.0$ (figure \ref{fig:competency_progress}).\\

\section{Discussion}
In these user studies, the overall progress of ranking correlations is measured as a metric to evaluate the user’s mental model of system competency. We test the user's mental model after they only see 100 instances (trials) of the AI agent's performance. However, the results strongly suggest that even with this limited view of system performance, the subjects learn the overall competency of AI agents throughout these tests.\\
Adding the attentional explanations for both models results in a significant improvement over competency rankings. Comparing the early learning rates between baseline and explanation groups suggests a significant improvement by attention map explanations especially for the SVQA model. However, the SVQA learning curve suggests an upper bound to the correlation in the presence of explanations. On the other hand, the SOBERT model shows a higher learning rate with explanations compared to the baseline while still reaching the maximum value of correlation.\\
These results highlight the effect of input features on the information that the explanations can carry. The SOBERT model uses object and spatial features vs. the spatial features in the SVQA model. The SOBERT model also uses BERT to transfer the features into attention maps. These changes w. r. t. the SVQA has raised the upper bound on the maximum reachable competency prediction by the subjects.\\
\section{Conclusion}
In this paper, we evaluate the role of attention map explanations on the user’s mental model of AI competency. We designed an experiment where subjects rank the performance of the VQA model among four different types of questions. To quantify the subjects’ mental model, we compute the correlation between user rankings and AI’s actual ranking among the question types.\\
We propose a new XVQA model that produces answers and attention maps from spatial and object features of the image. This explainable model uses a BERT language module to better process the visual and textual embeddings of the input. The proposed model is compared with a baseline model to show the effect of input object features and also the BERT attention module.\\
Overall results from the experiments suggest an improvement in the user mental model when exposed to the attention map explanations. The progress of the user mental model (ranking correlations) throughout the experiments indicates a higher learning rate in the presence of explanations. Furthermore, the subject group interacting with the newly proposed model shows a higher rate of ranking correlation compared to the baseline model. This improvement suggests a positive impact on the explanations by including the object feature and the BERT language model.\\
\FloatBarrier
{\small
\bibliographystyle{IEEEtran}
\bibliography{egbib}
}

\end{document}